\documentclass[conference]{IEEEtran}
\usepackage{amsmath, bm, bbm}
\usepackage{amssymb}
\usepackage{amsfonts}
\usepackage{graphicx}
\usepackage{subcaption}
\usepackage{mwe}
\usepackage{float}
\usepackage{color}
\usepackage{mathtools}
\usepackage{comment}

\usepackage{lipsum}% http://ctan.org/pkg/lipsum
\usepackage{multicol}% http://ctan.org/pkg/multicols
\usepackage{multirow}
\usepackage{graphicx}

\DeclarePairedDelimiter\floor{\lfloor}{\rfloor}
\newcommand{\norm}[1]{\left\lVert#1\right\rVert}

\newcommand{\etal}{\textit{et al}.}

\begin{document}

\title{Forecast-based Multi-aspect Framework for Multivariate Time-series Anomaly Detection}

\makeatletter
\newcommand{\linebreakand}{%
  \end{@IEEEauthorhalign}
  \hfill\mbox{}\par
  \mbox{}\hfill\begin{@IEEEauthorhalign}
}
\makeatother


\author{
    \IEEEauthorblockN{Lan Wang, Yusan Lin, Yuhang Wu, Huiyuan Chen, Fei Wang, Hao Yang}
    \IEEEauthorblockA{Visa Research, Palo Alto, CA, USA, 94306}
    \IEEEauthorblockA{\{lwang4, yusalin, yuhawu, hchen, feiwang, haoyang\}@visa.com}
}

%\IEEEoverridecommandlockouts
%\IEEEpubid{\makebox[\columnwidth]{978-1-6654-3902-2/21/\$31.00~\copyright2021 IEEE \hfill}
%\hspace{\columnsep}\makebox[\columnwidth]{ }}

\maketitle
 
%\IEEEpubidadjcol

%\begin{abstract}
%    Multivariate time series anomaly detection has been a long-standing research topic both in academic and industrial world. forecast-based approaches which detect anomalies based on prediction errors have been proved effective in the past. However, most existing models are one-fold regardless of the distinct natures of different types of anomalies, and thus resulting to either sub-optimal performance or performance inconsistency cross datasets. In this paper, we propose FMUAD: a forecast-based multi-aspect unsupervised anomaly detection framework, which explicitly addresses the characteristics of different anomaly types for multivariate time series: value change, frequency change and correlation change. We design a comprehensive architecture with three independent modules to capture anomaly pattern from different aspects, and then learn jointly to find the optimal feature representation. Unlike other deep learning models, our design is highly flexible, intuitive and explainable, which makes easy human interaction. Extensive experiments show that our framework outperforms all the other state-of-the-art forecast-based anomaly detection models.
%\end{abstract}

\begin{abstract}
    Today's cyber-world is vastly multivariate. Metrics collected at extreme varieties demand multivariate algorithms to properly detect anomalies. However, forecast-based algorithms, as widely proven approaches, often perform sub-optimally or inconsistently across datasets. A key common issue is they strive to be one-size-fits-all but anomalies are distinctive in nature. We propose a method that tailors to such distinction. Presenting FMUAD - a Forecast-based, Multi-aspect, Unsupervised Anomaly Detection framework. FMUAD explicitly and separately captures the signature traits of anomaly types - spatial change, temporal change and correlation change - with independent modules. The modules then jointly learn an optimal feature representation, which is highly flexible and intuitive, unlike most other models in the category. Extensive experiments show our FMUAD framework consistently outperforms other state-of-the-art forecast-based anomaly detectors.
\end{abstract}


\IEEEpeerreviewmaketitle

\section{Introduction}

Anomaly detection for multivariate time series is of great interest in many real-word applications, such as road traffic surveillance~\cite{wei2018unsupervised}, financial fraud detection~\cite{ahmed2016survey}, software monitoring \cite{Ghuli2018}, web log analysis \cite{Cao2017} and network analysis~\cite{ding2019interactive}, etc. Detecting unexpected behavior that deviates from the normal behavior accurately and efficiently can avoid further damage of core systems and save huge amount of revenue for business. 

Among the various challenges to the said anomaly detection application, the most important one lies in the nature of the multi-variate time series themselves. For example, anomalous events are the mutual effect of multiple time series, such that looking at each one separately can result in poor quality detection \cite{Zhao2020}. Intra-series patterns are far from trivial as well. The temporal dynamics, such as shift in frequency and abrupt change-of-trend \cite{Gao2020}, need specific modeling to make accurate forecasting based on historical data \cite{Cao2021}. In addition, value changes themselves may be subtle enough to disguise as normal data, but looking at the same time series in different granularities or scales tells completely different stories. All of these peculiarities need to be accounted for designing an effective detector.

\begin{figure*}
    \centering
    \begin{subfigure}[b]{0.35\textwidth}
        \centering
        \includegraphics[width=\textwidth]{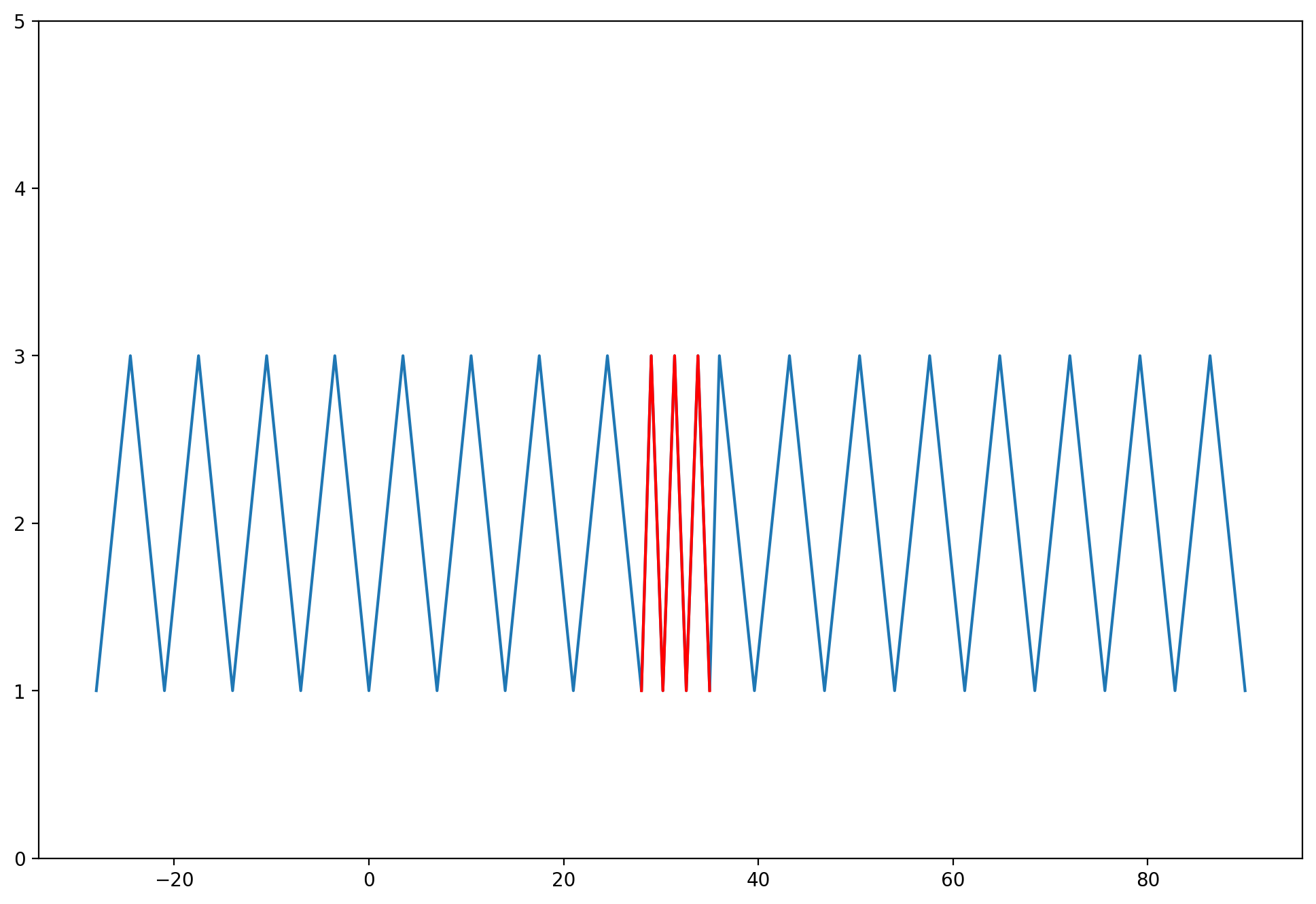}
        \caption[Network2]%
        {{\small Frequency change}}    
        \label{fig:ex1}
    \end{subfigure}
    %\hfill
    \begin{subfigure}[b]{0.35\textwidth}  
        \centering 
        \includegraphics[width=\textwidth]{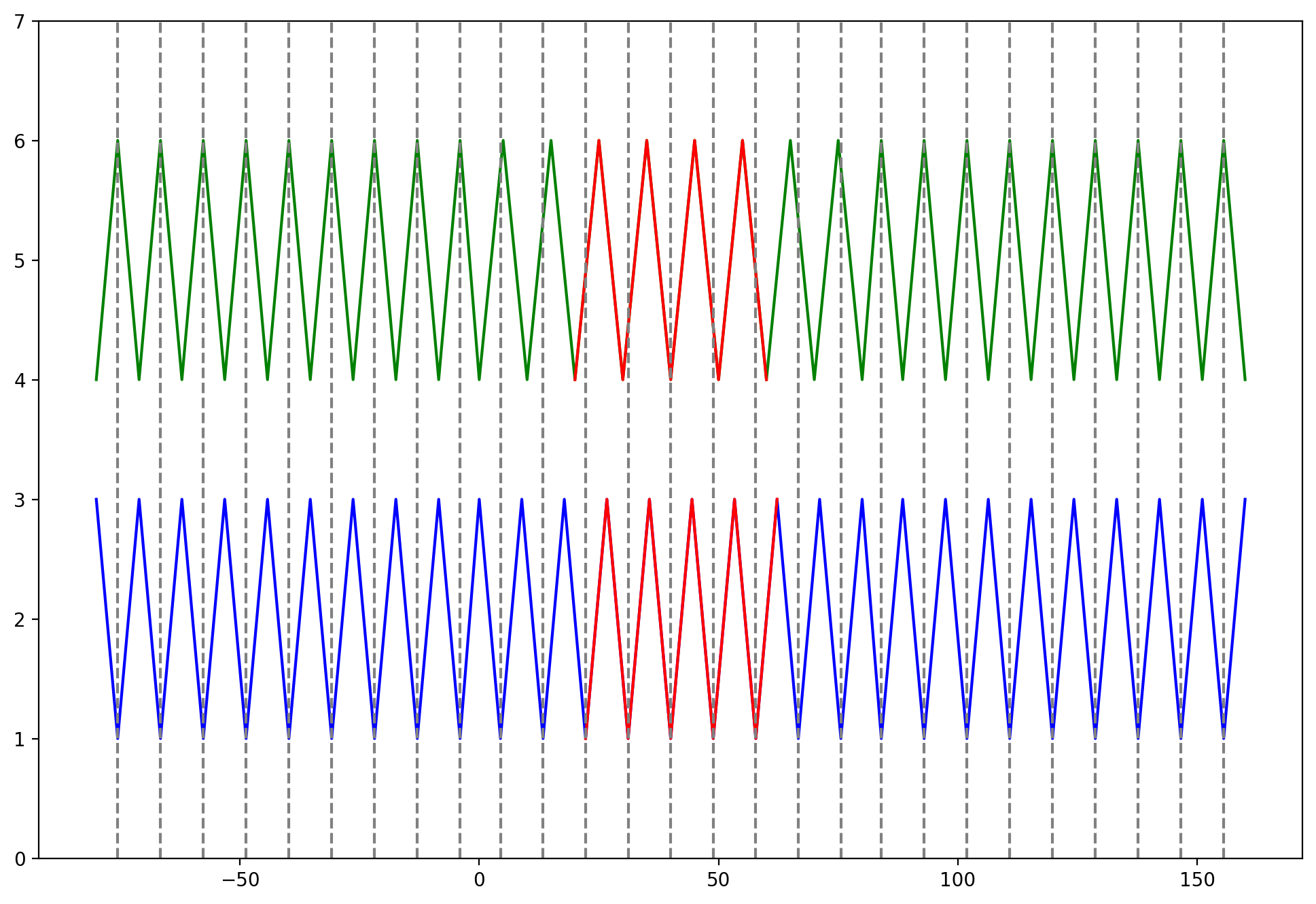}
        \caption[]%
        {{\small Correlation change}}    
        \label{fig:ex2}
    \end{subfigure}
    \vskip\baselineskip
    \begin{subfigure}[b]{0.35\textwidth}   
        \centering 
        \includegraphics[width=\textwidth]{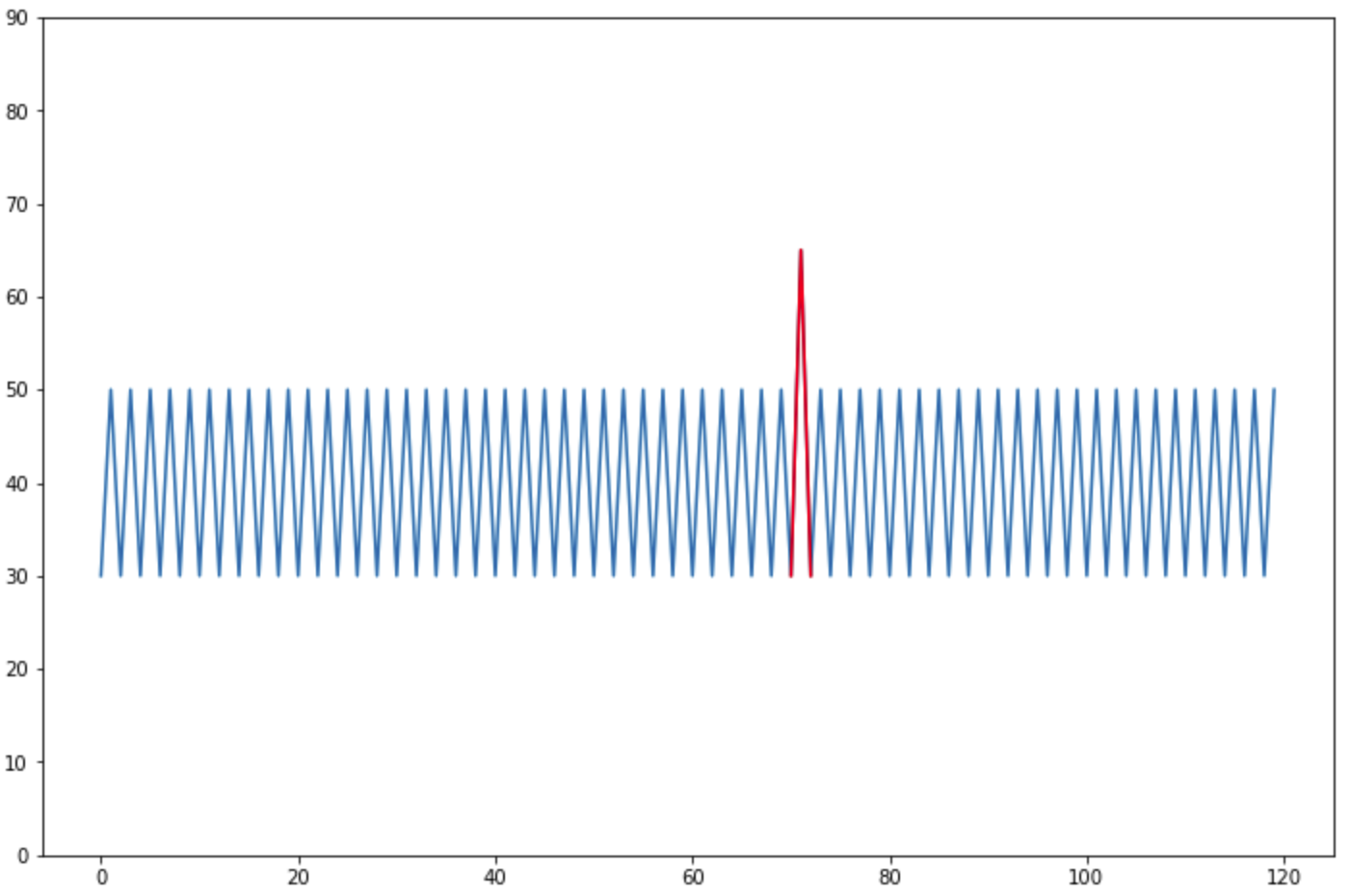}
        \caption[]%
        {{\small Value abrupt change}}    
        \label{fig:ex3}
    \end{subfigure}
    %\hfill
    \begin{subfigure}[b]{0.35\textwidth}   
        \centering 
        \includegraphics[width=\textwidth]{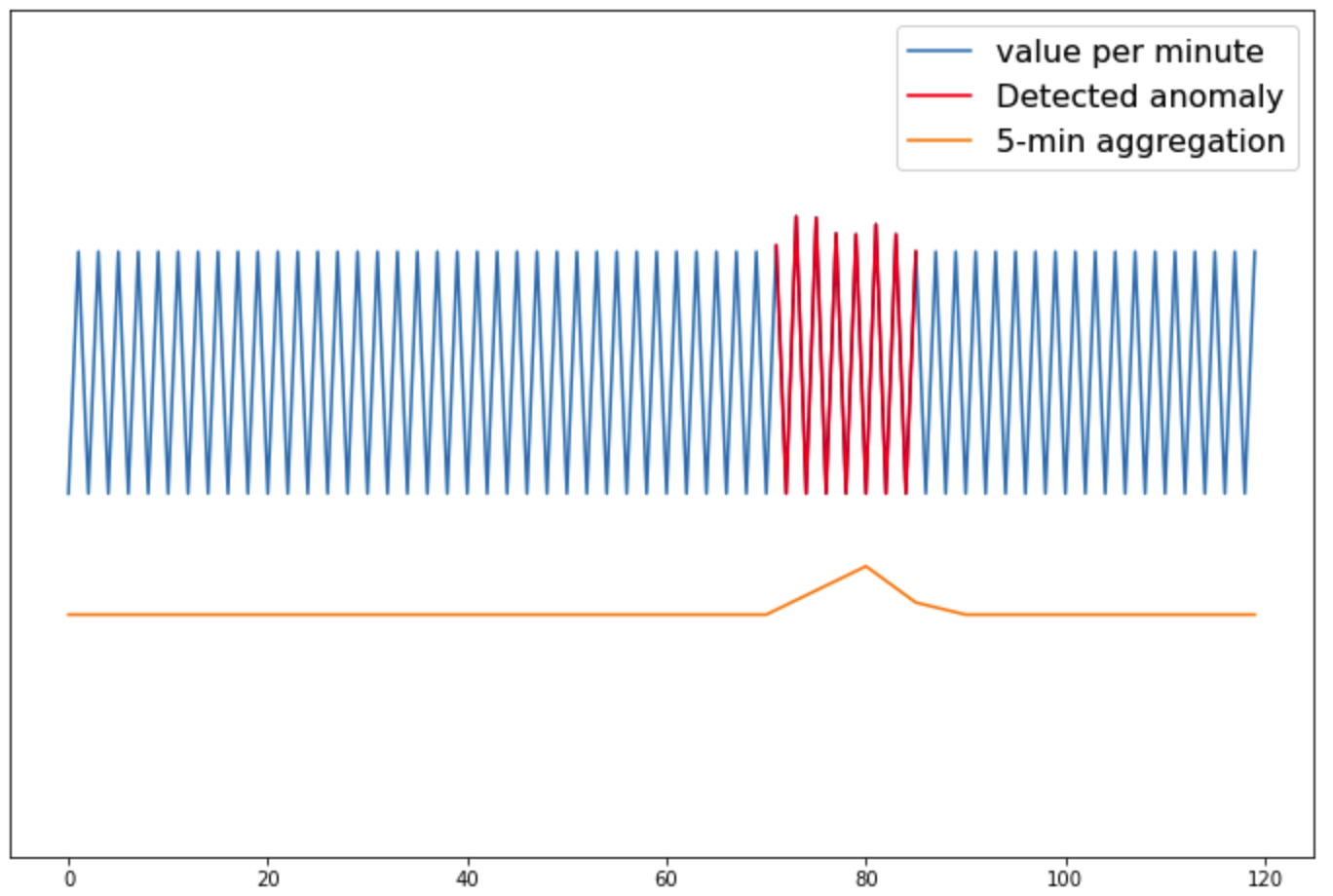}
        \caption[]%
        {{\small Value subtle change}}    
        \label{fig:ex4}
    \end{subfigure}
    \caption[Motivation]
    {\small Examples of anomaly patterns: (a) temporal change; (b) correlation change; (c, d) spatial change where (c) corresponds to a dramatic change and (d) more subtle change in which anomalies accumulates over a time period. The red region represents where the anomalies happen while data points located elsewhere are normal.} 
    \label{fig:exs}
\end{figure*}

Most forecast-based models in the past do not explicitly address the above-mentioned patterns, let alone separately \cite{Ghuli2018, DBLP:journals/corr/abs-2101-03064, Wu2020, zong2018deep, YiKe2018}, and therefore are subject to each of the above deficiencies. However, \textit{we argue that there is a way to fix such deficiencies and forecast-based models in general have room for improvement.} In this paper, we propose a modular approach whereby we create different modules to best characterize each of the three anomalous traits: inter-series correlation dynamics, intra-series temporal dynamics and  multi-scale spatial dynamics. These modules are trained jointly to provide a unified inference.

Another key concern in our proposal is the lack of labels for anomalous events~\cite{chalapathy2019deep}. As anomalies are commonly outnumbered by normal data points, it is crucial to have a label-agnostic model. Consequently, a noticeable trend in the domain is pivoting towards unsupervised models, i.e. ones that train on normal data exclusively, and detect anomalies with the intuition that they deviate from the patterns of normal events. Our proposal follows suit with this trend.

The main contributions of this paper are summarized as follows:

\begin{itemize}
    \item We introduce a Forecast-based Multi-aspect Unsupervised Anomaly Detection framework (FMUAD) to capture anomalies with different patterns. The model outperforms other state-of-the-art forecast-based methods by achieving over 17\% F1-score improvement on average cross multiple public datasets.
    
    \item We design a modular architecture that explicitly addresses different patterns of anomalies, which makes our framework highly intuitive and explainable. By learning different modules jointly, our approach not only captures each individual anomaly pattern but also detects complicated real-world anomalies with mixed patterns. 
    
    \item We define a novel loss function that involves the forecasting error as well as its compactness (or variance). Compactness has proved to be a desired quality of a feature to control the variance within one class \cite{DBLP:journals/corr/abs-2101-03064}. By controlling the compactness of normal class, we aim to learn a tighter representation that is more sensitive to outliers. 
\end{itemize}

\section{Related Work}

As Chandola \etal \cite{Chandola2009} put in his nice words, ``Anomalies are patterns in data that do not conform to a well defined notion of normal behavior." And such is the fundamental principle of unsupervised anomaly detectors (UADs) - Instead of learning to tell apart normal and abnormal events, UADs simply discover data that deviates from normal. Technically speaking, UADs leverage the following major characteristics of anomalies: 
(1) The distribution of anomalies differs remarkably from the normal data points;
(2) Anomalies happen infrequently and thus form a small set of data points compared with the normal data.

Following that, the importance of modeling normal behavior goes without saying. In particular, we need to find an effective way to represent normal data. Time series, a prime subject for anomaly detection, is often diverse and complex due to temporal dependency between data points. Some researchers attempted to leverage a graph structure to model the dynamic change \cite{Zhao2020, Cao2021, Spadon2021, DBLP:journals/corr/abs-2108-06783, 10.1145/3404835.3462868, chen2018exploiting}. However, graph construction and modeling adds extra complexity to the system, which makes it heavy and time costly, and thus are not scalable for real-world applications.  To this day, how to find an universally satisfactory representation for time series is still under intense discussions in the field. 

Among the vast literature on time series anomaly detection over the past decade \cite{Zhang2019,Su2019,Audibert2020,Gao2020,Zhao2021,Lee2021,Zhao2020}, one can divide them into three groups: (1) proximity-based; (2) reconstruction-based; and (3) forecast-based. Proximity-based methods quantify object similarities based on a defined distance measure and detect objects that are far away from the majority as anomalies. Reconstruction-based methods use the reconstruction error as the core and assume that anomalies lie in a different manifold from most data points, and thus cannot be effectively reconstructed from low-dimensional space. Forecast-based methods assume anomalies incorporate unusual patterns that are hard to predict based on historical data, and detect anomalies based on the forecasting error. Our work falls under this category. Below, we present a brief background, along with some important past work on forecast-based time series anomaly detection.

In a forecast-based anomaly detector, we prepare the model by fitting it against the training data, which only contains normal data points. The trained model is then put forth to make predictions on future ``normal" data, to be compared against the input. We infer the input point as anomalous if its value deviates from the model's prediction range. Given such fundamental mechanism, it easily follows that forecast-based anomaly detectors mainly differentiate by their forecasting methods. One of the most classic approaches, ARIMA \cite{7814437} is a popular statistical analysis model that learns the auto-correlations in the time series for future value prediction. Other statistics models such as Holt-Winters\cite{7814437} and FDA \cite{Torres2011} serve the same purpose. While these methods are efficient, they are sensitive to datasets and model parameter choices. Fine tuning these models often requires strong assumptions and extensive domain knowledge about the data. Machine-learning based models attempt to overcome such limitations. Hierarchical Temporal Memory (HTM) \cite{Ahmad2017} is an unsupervised sequence memory algorithm to detect anomalies in streaming data. Ding \etal \cite{Ding2018} combined the HTM model and Bayesian network into for a real-time framework for multi-variate time series anomaly detection. More recently, Long Short-Term Memory Recurrent Neural Network (LSTM-RNN), an effective deep learning network that models temporal dynamics for sequential data, has been widely used for many state-of-the-art works. Hundman \etal \cite{Hundman2018} proposed LSTM-NDT, an unsupervised and non-parametric thresholding approach to interpret prediction generated by an LSTM network. Zong \etal \cite{zong2018deep} proposed DAGMM to detect anomalies by jointly optimizing the parameters of a deep neural network and a Gaussian Mixture Model.

Although the past forecast-based models have proven the category's potential in anomaly detection, they were designed to be a capture-all solution for every possible anomaly, which can be risky due to the drastically difference between the anomaly patterns.  We believe a degree of tailoring in the model to anomaly patterns can help bring out better performance. 

% None of them explicitly address anomalies with different patterns that appears to be the case especially for multi-dimensional data, but rather focus on a one-size-fits-all module in hope to capture all. In this paper, we design a modular framework that shows superiority over existing state-of-the-art works.

\section{The Proposed Method}

We propose an unsupervised, forecast-based, multi-variate time-series anomaly detection algorithm (FMUAD) to explicitly address different natures of anomaly patterns - temporal change, spatial change and correlation change, which are challenging to detect using forecast-based detectors of the past. In this section, we first formalize our problem and then provide an overview of the model methodology.

\subsection{Problem Formulation}
A multivariate time series is a sequence of multi-dimensional data points 
\[
    \mathcal{X} = \{\bm{x}_1, \bm{x}_2, ...., \bm{x}_T\},
\]
where $\bm{x}_i \in \mathbb{R}^{m}$ for $i\in\{1, ..., T\}$, $m$ is the number of features and $T$ is the number of time steps.  The uni-variate setting is a particular case of the multi-variate one with $m=1$.

The model takes $\mathcal{X}$ as the input training data assuming $\mathcal{X}$ contains only normal time steps. Anomaly detection refers to identifying a new unobserved instance $\hat{\bm{x}}_t, t>T$ based on how differently it behaves compared to $\mathcal{X}$. Due to the temporal dependency within the time series and the fact that anomalies usually occur over a contiguous time period, we do not model each time step independently but rather consider a window segment. Suppose $W_t \in \mathbb{R}^{m\times k}$ is a window with length $k$ at time $t$:
\[
W_t = \{\bm{x}_{t-k+1}, \bm{x}_{t-k+2}, ..., \bm{x}_t\}.
\]

We can transform the original time series $\mathcal{X}$ into a sequence of windows $\bm{W} = \{W_{k}, ..., W_T\}$. For each window $W_t$, we train the model to forecast its behavior using the historical time steps that were already seen. With proper training, the model will be capable of making decent predictions for normal data. Then given a new unseen window $\tilde{W}_t, t>T$, our goal for the model is to detect whether an anomaly happens at time $t$ based on the forecast error: the larger the error is, the more likely an anomaly happens.\\

%We can transform the original time series $\mathcal{X}$ into a sequence of windows $\bm{W} = \{W_{k}, ..., W_T\}$. For each window $W_t$ seen in the data, we aim to both derive the behavior of $W_t$ (denoted as $Y_t$ thereafter), as well as forecast the normal behavior at time $t$ based on past input data $W_{...t}$ (denoted $\hat_Y_t$). The anomaly detection is achieved by judging the deviation of the derived behavior $Y_t$ and forecasted behavior $\hat{Y}_t$. Specifically, we output $\hat{y}_t = 1$ if the forecast error is big enough to say it's an anomaly, and $\hat{y}_t = 0$ otherwise.

% We transform the input data $\mathcal{X}$ into windows $\{W_{k}, W_{k+1}, ..., W_T\}$. By learning the representation of these normal windows, we hope to correctly assign a binary label $y\in\{0,1\}$ given an unobserved window based on its anomaly score, where $y=1$ refers to anomaly and $y=0$ a normal window. 

\begin{table}[ht]
    \centering
    \caption{Notations}
    \begin{tabular}{cc}
         \hline
         % TODO: W_t, I_t, Y_t, y_t
         $m\in\mathbb{Z}$ & Number of time series (features) \\
         $T\in\mathbb{Z}$ & Number of total time steps \\
         $k\in\mathbb{Z}$ & Window size \\
         $\tau\in\mathbb{Z}$ & Input instance time span\\
         $\mathcal{X}\in\mathbb{R}^{m\times T}$ & Input time series\\
         $I_t\in\mathbb{R}^{m\times \tau}$ & Individual input instance\\
         $W_t\in\mathbb{R}^{m\times k}$ & True window at time $t$\\
         $S_t\in\mathbb{R}^{m\times m}$ & True signature matrix (correlation) for $W_t$\\
         $F_t\in\mathbb{R}^{m\times k/2}$ & True frequency matrix for $W_t$ \\
         $W^h_t\in\mathbb{R}^{m\times (\tau-k)}$ & Historical segment for $W_t$\\
         $\hat{W}_t\in\mathbb{R}^{m\times k}$ & Predicted window at time $t$\\
         $\hat{S}_t\in\mathbb{R}^{m\times m}$ & Predicted signature matrix (correlation) for $W_t$\\
         $\hat{F}_t\in\mathbb{R}^{m\times k/2}$ & Predicted frequency matrix for $W_t$ \\
         \hline
    \end{tabular}
    \label{tab:notations}
\end{table}

\subsection{Our Method}

In this section, we first present the overall architecture of the model and then break down each module into details.

\subsubsection{Overall Architecture}

%As depicted in Figure \ref{fig:arch}, the input data to our framework follows the form of a $m\times \tau$ matrix:

We illustrate the overall architecture of our model in this part. See Figure \ref{fig:arch} for a high-level walk-through and Figure \ref{fig:network} for a more detailed illustration. 

To detect anomaly in a target window $W_t$, we concatenate the necessary historical information needed for forecasting, or $W^h_t$, with $W_t$ itself, to formulate the input to our model:
\[
I_t=[W^h_t; W_t],
\]
which takes the form of a $m\times \tau$ matrix:
\[
I_t = \{\bm{x}_{t-\tau+1}, \bm{x}_{t-\tau+2}, ..., \bm{x}_t\},
\]
where $m$ is the number of features and $\tau$ is the number of time steps we trace back for each prediction. In our setting, we choose $\tau=500$ and $k=30$. The output of our framework, $\hat{y}_t \in \{0,1\}$ is a boolean representing whether anomaly happens at time $t$.

\begin{figure}[H]
    \centering
    \includegraphics[width=80mm]{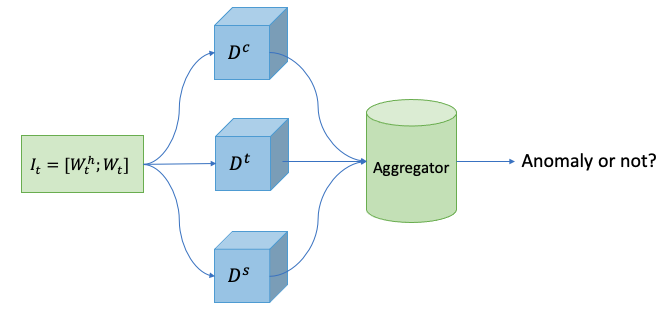}
    \caption{High-level system architecture}
    \label{fig:arch}
\end{figure}

To holistically characterize and capture the traits of the anomalies within each window, we consider the time series dynamics from three aspects and design a corresponding detector tailored to each aspect:

\begin{itemize}
    \item[--] $D^c$, a detector for \textbf{inter-series correlation} change;
    \item[--] $D^t$, a detector for \textbf{intra-series temporal pattern} change;
    \item[--] $D^s$, a detector for \textbf{multi-scale spatial pattern} change.
\end{itemize}

Because the nature of the detectors are largely dictated by the traits they're assigned to capture, each detector should use a different input/output data representation that best suits the detector. To this regard, we create two transformed matrices $S_t$ and $F_t$, and use them along with $W_t$ for the three detectors $D^c, D^t, D^s$ respectively. The transformation process TF1: $ W_t \mapsto S_t$ and TF2: $ W_t \mapsto F_t$ will be discussed in details in part 2) and 3).

\begin{itemize}
    \item[--] $W_t$, the original matrix for the window at $t$;
     \item[--] $F_t$ (frequency matrix), a derived matrix that contains temporal information for $W_t$ in the frequency domain;
    \item[--] $S_t$ (signature matrix), a derived matrix that contains inter-series correlation information for $W_t$.
\end{itemize}

\begin{figure}[H]
    \centering
    \includegraphics[width=60mm]{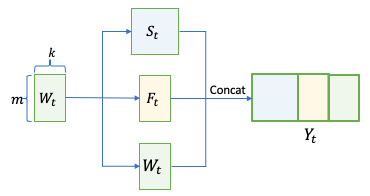}
    \caption{Concatenation of three matrices}
    \label{fig:concat}
\end{figure}
As depicted by Figure \ref{fig:concat}, $Y_t$ is a concatenation of the ground truth matrices $S_t, F_t$ and $W_t$, i.e.,
\[Y_t = [S_t; F_t; W_t] = [\text{TF1}(W_t); \text{TF2}(W_t); W_t].\]
% $$\begin{aligned}
% Y_t &= [S_t, F_t, W_t] \\
%     &= [\text{TF1}(W_t), \text{TF2}(W_t), W_t]
% \end{aligned}$$
Our model outputs $\hat{Y}_t = [\hat{S}_t; \hat{F}_t; \hat{W}_t]$ as the predicted matrix derived from the concatenation of the forecasting results obtained from the three detectors. The error between $Y_t$ and $\hat{Y}_t$ will be served as an indicator of anomalies. 

%Each of the three detectors use $W^h_t$ to forecast a normal reference behavior  $\hat{Y}_t = [\hat{S}_t; \hat{F}_t; \hat{W}_t]$. (Figure \ref{fig:network} details the full process) We compute the error between $Y_t$ and $\hat{Y}_t$, which will be served as an indicator of anomalies. 

%The final predicted matrix is derived from the aggregation of the forecasting results obtained from the three detectors. With the final target and prediction, we finally assign a binary label indicating anomalies at time $t$ based on the forecasting error: the higher the error is, the more likely an anomaly happens. See Figure \ref{fig:network} for more detailed walk-flow.

\begin{figure*}[ht]
    \centering
    \includegraphics[width=180mm]{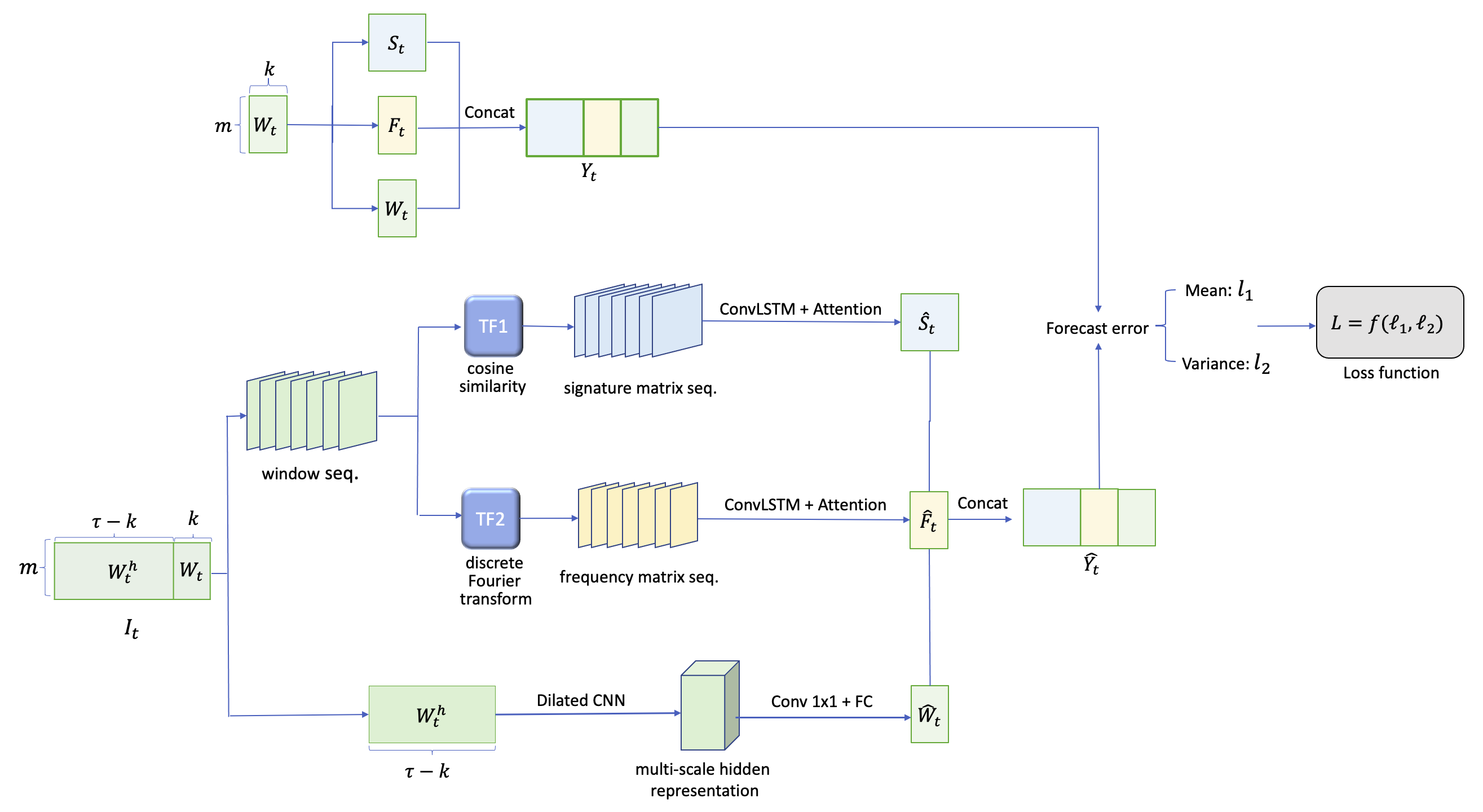}
    \caption{Detailed model architecture}
    \label{fig:network}
\end{figure*}

\subsubsection{\textbf{Inter-series correlation detector}}
Previous studies \cite{Zhang2019, 10.1145/3219819.3220108} suggest the importance of correlation among time series in the characterization of the dynamic of the multi-variate time series system for anomaly detection. For instance, the anomaly shown in the red region of Figure \ref{fig:exs}(b) is determined based on abnormal correlation: The two time series have been mostly negative correlated, while in the labelled pattern they are positively correlated.

To capture the correlation changes, inspired by \cite{Zhang2019}, we define a signature matrix $S_t$ to characterize the correlation between time series. The transformer TF1 that produces $S_t$ is defined as follows: Given a window $W_t$, $S_t$ is a $m\times m$ matrix whose entry at the $i$th row and $j$th column is computed as the Cosine similarity score between the $i$th time series $\{x^i_{t-k+1}, ... , x^i_{t}\}$ and the $j$th time series $\{x^j_{t-k+1}, ... , x^j_{t}\}$. The reason we adopt Cosine similarity instead of the inner product as used in \cite{Zhang2019} is that the former is less prone to scaling effects, and allows us to focus on the correlation effects only.

Now we transform the input time series into a sequence of signature matrices. First, we splice the input $I_t$ into a sequence of windows with stride $s$. Then, we apply TF1 to each of the window to form a sequence containing $d=\floor{(\tau-k)/s}$ signature matrices.
\begin{equation}
    {S}_t  	\leftarrow \text{TF1} (W_t).
    \label{DFT}
\end{equation}
To make a forecast on the signature matrix $S_t$, we apply ConvLSTM with an attention mechanism. The final forecasting signature matrix is denoted as $\hat{S}_t \in\mathbb{R}^{m\times m}$. ConvLSTM \cite{10.5555/2969239.2969329} has been developed and proven effective to capture the sequential temporal information, as formulated below:
\begin{equation}
    \begin{aligned}
    i_t =& \sigma(W_{si}*{S}_t + W_{hi}*{H}_{t-1} + W_{ci}\circ{C}_{t-1} + b_i) \\
    f_t =& \sigma(W_{sf}*{S}_t + W_{hf}*{H}_{t-1} + W_{cf}\circ{C}_{t-1} + b_f) \\
    {C}_t =& f_t\circ{C}_{t-1} + i_t\circ\tanh{(W_{sc}*{S}_t + W_{hc}*{H}_{t-1} + b_c)} \\
    o_t =& \sigma(W_{so}*{S}_t + W_{ho}*{H}_{t-1} + W_{co}\circ{C}_t + b_o)\\
    {H}_t =& o_t\circ\tanh({C}_t),
\end{aligned}
\label{eq:convlstm}
\end{equation}
where $i_t$, $f_t$, and $o_t$ denote the input gate, forget gate, and output gate, respectively. $C_t$ and $H_t$ are the cell states and hidden states. $*$ denotes the convolution operator, $\circ$ is the Hadamard product, and $\sigma(\cdot)$ is the sigmoid function. 

In addition, since not all previous hidden states have equal effects on the current state, we apply a temporal attention mechanism to adaptively adjust the weight of the previous steps that are relevant to the current one and aggregate all to obtain the overall effect~\cite{NIPS2017_attention}. More concretely, we have
\begin{equation}
     {H}^*_t = \sum_{i=t-k}^{t} c_i {H}_i, \quad 
     c_i = \frac{\exp{\langle \text{vec}(H_i), \text{vec}(H_t)\rangle}}{\sum_{j=t-k}^{t} \exp{\langle\text{vec}(H_j), \text{vec}(H_t)\rangle}},
     \label{eq:attention}
\end{equation}
where $\text{vec}(\cdot)$ is the matrix flattened into a vector, $\langle \cdot,\cdot\rangle$ denotes the inner product, and ${H}^*_t$ is the representation for window $t$. Then with an additional convolutional neural network with a $1\times 1$ kernel, we project the representation ${H}^*_t$ into a lower dimensional space with only one channel and then apply a fully-connected layer to derive the prediction of $S_t$:
\begin{equation}
    \hat{S}_t 	\leftarrow  \text{FC}(\text{Conv}_{1 \times 1}({H}^*_t )).
    \label{eq:S_update}
\end{equation}
The output serves as the final prediction $\hat{S}_t$ containing forecasted correlation information for window at time $t$.

\subsubsection{\textbf{Intra-series temporal pattern detector}}
As are inter-series correlations, the intra-series temporal patterns are also crucial for anomaly detection. For instance, the anomaly shown in the red region of Figure \ref{fig:exs}(a) is due to temporal pattern changes: the frequency suddenly increases twice in the anomaly region compared to the rest of the time series.  Several recent work \cite{Cao2021, Spadon2021} has shown the advantages of joint learning inter-series correlations and intra-series temporal patterns for multi-variate time series forecasting.

To capture the temporal pattern within each series, we project each uni-variate time series to the spectral domain via DFT (discrete Fourier transform):
\begin{equation}
    \xi_j = \frac{1}{k}\sum\limits_{\ell=0}^{k-1}x_l e^{2\pi i j\ell/k}, j=1,2,...,k,
\end{equation}
where $i$ is the imaginary unit. The DFT transforms $k$ discrete-time samples $\Vec{x} = \{x_1, x_2, ..., x_k\}$
in the time domain to the frequency domain $\Vec{\xi} = \{\xi_1, \xi_2, ..., \xi_k\}$. Due to the symmetry property of DFT for real-valued discrete-time signals: $\xi_j = \xi_{k-j}$, only half of the $\xi$'s need to be stored. As a result, the window $W_t$ can be transformed to a frequency matrix $F_t = \{\xi_1, ...,\xi_{k/2}\} \in\mathbb{R}^{m\times\frac{k}{2}}$ representing the amplitude of frequencies for the $m$ uni-variate time series in the spectral domain. Such process defines the transformer TF2.
\begin{equation}
    {F}_t  	\leftarrow \text{TF2} (W_t).
    \label{DFT}
\end{equation}

Similar to how the inter-series correlation was constructed in part 2), we transform the input time series into a sequence of frequency matrices, then apply ConvLSTM with an attention mechanism to obtain a high-dimensional hidden representation. Lastly we pass an additional neural network with a $1\times 1$ kernel to project it to a low-dimensional space and apply a fully-connected layer:
\begin{equation}
    \hat{F}_t   \leftarrow \text{FC}(\text{Conv}_{1 \times 1}(\text{Attn} (\text{ConvLSTM} (F_{t^-})))).
    \label{F_label}
\end{equation}
Here, $\text{ConvLSTM}(\cdot)$ refers to Eq. (\ref{eq:convlstm}),  $\text{Attn}(\cdot)$ refers to Eq. (\ref{eq:attention}), and $F_{t^-}$ refers to the sequence of frequency matrices before time $t$. The forecast $\hat{F}_t$ captures the temporal information from each time series within window $W_t$.

\subsubsection{\textbf{Multi-scale spatial pattern detector}}

Another pattern of anomaly, sometimes overlooked but extremely common in time series is the spacial dynamic, or in layman's term, ``value change". For instance, the red region of Figure \ref{fig:exs}(c) shows a dramatic value change, which is trivial to capture. However, the majority of spatial changes are gradual and subtle, and hard to detect like the red region in Figure \ref{fig:exs}(d). The value change at each individual time step is so small that they are hardly discernible to the human eye, but when accumulated over a time period, the deviation is significant. Thus, it is important for our anomaly detector to capture both types of spacial dynamics.

% The third major part of our design focuses on the spatial dynamic for the input time series. For instance, the red region of Figure \ref{fig:exs}(c) shows a dramatic value change. Such change is trivial and easy to capture. However, in real-world problems, spatial changes could be subtle and hard to detect like the red region in Figure \ref{fig:exs}(d). The value change at each individual time step is small, but the accumulated deviation is significant. Thus, it is important for our anomaly detector to capture both types of errors.

The intuition is natural: Aggregating the time series in Figure \ref{fig:exs}(d) at a larger time scale easily manifests the subtle changes as a spike, thus reducing the problem to a trivial one. In fact, looking at time series at different levels of aggregation can tell completely different stories. To capture multi-scale patterns, one needs to consider relatively long-term dynamics, which is why we take $W^h_t\in\mathbb{R}^{m\times(\tau-k)}$ to model instead of taking individual windows as we did before. In our model, we adopt the Dilated Convolutional Neural Network \cite{Yu2016MultiScaleCA} for this purpose. The major difference between dilated CNN and basic CNN is that it introduces dilated convolution operator to capture information aggregated at multiple scales. Suppose $F$ is a discrete function and $k$ is the convolution kernel, we can define $\ell$ as a dilation factor and $*_\ell$ as a dilated convolution operator
\begin{equation}
    (F *_\ell k)(p) = \sum\limits_{s+\ell t=p} F(s)k(t).
\end{equation}
With different dilation sizes $\ell$, multi-scale information can be obtained without losing resolution.

In our setting specifically, we firstly apply a 3-layer Dilated-CNN with dilating numbers $r=1,3,5$ and respective channels numbers $c=32,64,128$ to obtain a deep hidden representation. In particular, we apply a rectangle kernel $(3, 1)$ for each layer instead of an usual square kernel $(3, 3)$ to limit the multi-scale aggregation to over the time dimension rather than the feature dimension. Then we decode the hidden representation to forecast the target window $W_t$. The final forecasting matrix is denoted as $\hat{W}_t \in\mathbb{R}^{m\times k}$:
\begin{equation}
    \hat{W}_t  \leftarrow 	\text{FC}(\text{Conv}_{1\times 1}(\text{Dilated-CNN} (W^h_t))).
    \label{W_update}
\end{equation}

%In our setting specifically, we firstly apply a 3-layer Dilated-CNN with dilating numbers $r=1,3,5$ and respective channels numbers $c=32,64,128$ to obtain a deep hidden representation. In particular, we apply a rectangle kernel $(3, 1)$ instead of a normal square kernel $(3, 3)$ to limit the multi-scale aggregation to over the time dimension rather than the feature dimension. Then we pass the hidden representation into another convolutional layer with a $1\times 1$ filter to project to a representation with a single channel; lastly we apply a fully-connected layer to forecast the target window $W_t$ with an appropriate reshape operator. The final forecasting matrix is denoted as $\hat{W}_t \in\mathbb{R}^{m\times k}$. 

\begin{table*}[ht]
\centering
\begin{tabular}{|c|c|c|c|c|c|c|c|c|c|}
\hline
\multirow{2}{*}{Method} & \multicolumn{3}{c|}{SMD}             & \multicolumn{3}{c|}{MSL}             & \multicolumn{3}{c|}{Total}           \\ \cline{2-10} 
                        & Precision & Recall & F1              & Precision & Recall & F1              & Precision & Recall & F1              \\ \hline
LSTM-NDT                & 0.5684    & 0.6438 & 0.6037          & 0.5934    & 0.5374 & 0.5640          & 0.5809    & 0.5906 & 0.5839          \\ \hline
DAGMM                   & 0.5951    & 0.8782 & 0.7094          & 0.5412    & 0.9934 & 0.7007          & 0.5681    & 0.9358 & 0.7051          \\ \hline
LSTM-VAE                & 0.7922    & 0.7075 & 0.7842          & 0.5257    & 0.9546 & 0.6780          & 0.6590    & 0.8311 & 0.7311          \\ \hline
\textbf{FMUAD}          & 0.8574    & 0.8789 & \textbf{0.8680} & 0.8311    & 0.8786 & \textbf{0.8539} & 0.8443    & 0.8664 & \textbf{0.8610} \\ \hline
\end{tabular}
\caption{Performance of FMUAD and 3 baseline approaches.}
\label{tab:result}
\end{table*}

\subsubsection{Loss function}
In this section, we introduce a novel loss function based on the forecasting error as well as its variance.

Following steps 2)-4), for a given target window $W_t$, we have derived the ground truth matrix $Y_t=[S_t; F_t; W_t]$ and the forecasted matrix $\hat{Y}=[\hat{S}_t; \hat{F}_t; \hat{W}_t]$ as shown in Figure \ref{fig:network}. The forecasting error is the $L_2$ norm between $Y_t$ and $\hat{Y}_t$. As we train our neural network in batches of size $b$, the batch loss is averaged over the batch as shown in Eq. (\ref{loss1}). The total loss is further averaged across the batches.
%Derived from steps 2)-4), now given the target window $W_t$, we have three associated true characteristic matrices: $S_t, F_t, W_t$, and three predicted characteristic matrices: $\hat{S}_t, \hat{F}_t, \hat{W}_t$. Direct concatenation yields a forecasting input matrix $Y_t=[S_t, F_t, W_t]$ and a forecasting output matrix $\hat{Y}=[\hat{S}_t, \hat{F}_t, \hat{W}_t]$ as shown in Figure \ref{fig:network}. We train our neural network in batches and the total loss is averaged over all the batches. Suppose $b$ denotes the batch size, then the mean forecasting error within each batch is defined as the average of $L_2$ norm of the difference between $Y_t$ and $\hat{Y}_t$:
\begin{equation}
    \ell_1 = \frac{1}{b}\sum_{i=1}^b\norm{Y_i - \hat{Y}_i}^2_2.
    \label{loss1}
\end{equation}

Besides forecasting error, we introduce an extra term we call the \textit{compactness loss} to capture the variance of the forecasting error within each batch, which is defined in Eq. (\ref{loss2}).
\begin{equation}
    \ell_2 =  \frac{1}{nb}\sum_{i=1}^b z_i^Tz_i, \ z_i = \hat{Y}_i - \frac{1}{b-1}\sum_{j\neq i}\hat{Y}_j,
    \label{loss2}
\end{equation}
where $n = m+k+\frac{k}{2}$ is the number of columns of $\hat{Y}_i$. Due to the fact that we are training on a single class (normal instances), the variance control $\ell_2$ helps to obtain a compact representation that is more sensitive to anomalies \cite{DBLP:journals/corr/abs-2101-03064, Perera2019}. 

The final loss is computed based on $\ell_1$ and $\ell_2$ via a function $f$ as defined below:

\begin{equation}
    \ell = f(\ell_1, \ell_2) =(\epsilon + \ell_2) \cdot \ell_1.
    \label{loss_fun2}
\end{equation}
Here, we take a small positive constant $\epsilon$  (In our experiment, we set $\epsilon = 10^{-5}$) to avoid a zero trivial solution: A trivial network with all zero parameters gives zero loss value $\ell$ if $\epsilon=0$. We adopt this loss function instead of a weighted sum of $\ell_1$ and $\ell_2$ due to two reasons. First, this eliminates the need to fine-tune the weights, making the training process much more efficient. Second, the effect each term adds on scales with its order of change. In other words, if the mean and variance of the forecasting error have the same order of changes, then these changes are emphasized to the same level no matter how different their scales are. To the best of our knowledge, this is the first work that incorporates the product of the error mean and variance in the loss function. 

%Here, we take a small positive constant $\epsilon$ to avoid a zero trivial solution (In our experiment, we set $\epsilon = 10^{-5}$). Note that a trivial network with all zero parameters gives zero loss value $\ell$ if $\epsilon=0$. We adopt this new loss function instead of a simple weighted sum of $\ell_1$ and $\ell_2$ due to two reasons. First, there is no need of fine-tuning on hyper-parameters (weight), which makes the training process much more efficient. Second, the effect each term adds on scales with its order of change. In other words, if the mean and the variance of the forecasting error have the same order of changes, then these changes are emphasized to the same level no matter how different their scales are. 

% While we consider both the forecasting error and its variance in the training phase, in the testing phase, we only use the forecasting error $\|Y_t - \hat{Y}_t\|^2_2$ to compute the loss for a given new window of multi-variate time series. The larger the testing loss is, the more likely an anomaly occurs. Our experiments also show that by considering the variance factor $\ell_2$, our model performance can be consistently improved. 

Our experiments show that by considering the variance factor $\ell_2$, the model performance can be consistently improved. In addition, While we consider both the forecasting error and its variance in training, we only use the forecasting error $\ell_1 = \|Y_t - \hat{Y}_t\|^2_2$ as the testing loss. The testing loss serves as the anomaly score to make the judgement: the larger the testing loss is, the more likely an anomaly has occurred.

 \begin{figure*}[!htbp]
    \centering
    \includegraphics[width=150mm]{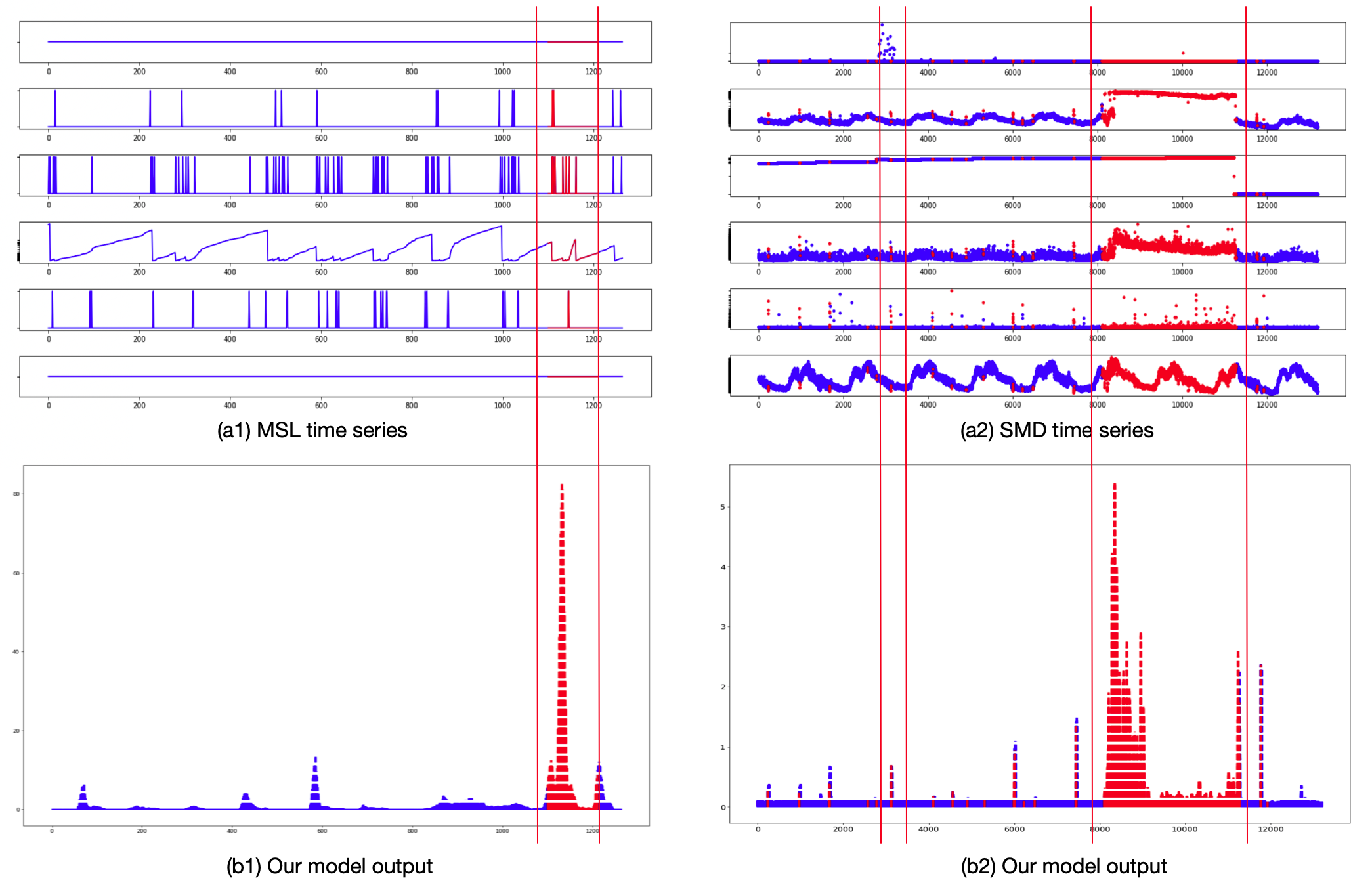}
    \caption{Case studies - (a1, b1): anomaly detection for MSL; (a2, b2): anomaly detection for SMD. (a1, a2) corresponds to the original time series and (b1, b2) corresponds to the anomaly scores output by our framework.}
    \label{fig:case_study}
\end{figure*}

\section{Experiments}
 
In this section, we perform extensive experiments to validate the effectiveness of our framework. We first describe the datasets that the framework tests on, then introduce three baseline models and the performance metrics, and make a comprehensive comparison between our model the baselines. Detailed case study and ablation study is presented at the end.

\subsection{Datasets}

To intuitively illustrate the inner workings of our model and evaluate its performance, we adopt two highly regarded and commonly used public multi-variate time-series datasets: SMD\footnote{\scriptsize https://github.com/NetManAIOps/OmniAnomaly/tree/master/ServerMachineDataset} (Server Machine Dataset) \cite{Su2019} and MSL\footnote{\scriptsize https://s3-us-west-2.amazonaws.com/telemanom/data.zip} (Mars Science Laboratory rover) \cite{Hundman2018}. SMD is 5 weeks worth of server telemetry collected from 28 production server nodes of a reputable Internet company. It contains 38 simultaneously collected floating-point metrics, with anomaly events labelled in a binary format and the anomaly ratio is 4.16\%. The data is segmented among the 28 servers into 3 groups. We train and test on each server machine separately. The total training set for SMD contains 708405 data points while the testing set contains 708420. The MSL dataset captures telemetry data throughout the Mars mission duration of the MSL rover, better known as \textit{Curiosity}. It contains data for 55 entities each monitored by 25 metrics, with anomalies expertly labelled by NASA and the anomaly ratio is 10.72\%. The training set size for MSL is 58317 while the testing size is 73729. Both datasets depict highly real-world scenarios in the target applications of our model, and therefore can provide an accurate benchmark for our model.

% generated by https://www.tablesgenerator.com/
\begin{table}[H]
\begin{tabular}{|cc|c|c|c|}
\hline
\multicolumn{2}{|c|}{Dataset Name}                   & \# Features         & Feature Name                                                                                                                & \multicolumn{1}{l|}{\begin{tabular}[c]{@{}l@{}}Anomaly \\ Ratio(\%)\end{tabular}} \\ \hline
\multicolumn{1}{|c|}{\multirow{3}{*}{SMD}} & group 1 & \multirow{3}{*}{38} & \multirow{3}{*}{\begin{tabular}[c]{@{}c@{}}CPU load, \\ network usage,\\ memory usage, etc.\end{tabular}}                   & \multirow{3}{*}{4.16}                                                             \\ \cline{2-2}
\multicolumn{1}{|c|}{}                     & group 2 &                     &                                                                                                                             &                                                                                   \\ \cline{2-2}
\multicolumn{1}{|c|}{}                     & group 3 &                     &                                                                                                                             &                                                                                   \\ \hline
\multicolumn{2}{|c|}{MSL}                            & 55                  & \begin{tabular}[c]{@{}c@{}}Telemetry data: \\ radiation,  power,\\ temperature, \\ computational activity, etc\end{tabular} & 10.72                                                                             \\ \hline
\end{tabular}
\caption{Dataset descriptions}
\end{table}

\begin{comment}
\begin{table}[H]
    \centering
    \caption{Datasets}
    \begin{tabular}{|c|c|c|c|c|c|}
         \hline 
         \textbf{Dataset} & \textbf{Number of entities} & \textbf{Number of features} & \textbf{Training set size} & \textbf{Testing set size} & \textbf{Anomaly ratio (\%)}  \\
         \hline
         SMD & 28 & 38 & 708405 & 708420 & 4.16 \\
         \hline
         SMAP & 55 & 25 & 135183 & 427617 & 13.13 \\
         \hline
         MSL & 27 & 55 & 58317 & 73729 & 10.72 \\
         \hline
    \end{tabular}
    \label{tab:datasets}
\end{table}
\caption{Descriptions of public datasets}
\end{comment}

 \subsection{Comparison with Baselines}

  %Hierarchical Temporal Memory (HTM) \cite{Ahmad2017},
 We compare our model with three state-of-the-art anomaly detection models: LSTM-NDT \cite{Hundman2018}, DAGMM \cite{zong2018deep} and LSTM-VAE \cite{DBLP:journals/corr/abs-1711-00614}. The first two models are forecast-based, which falls in the same scope as our model. The last one is also included here for comparison is due to the fact that they also adopted LSTM as a substantial part in their design to capture temporal dynamics. To ensure a fair comparison, we use the same set of evaluation metrics as the baselines: precision (P), recall (R) and F1 score (F1):
 \[
 P=\frac{TP}{TP+FP},\ R=\frac{TP}{TP+FN},\ F1=2\cdot\frac{P\cdot R}{P+R},
 \]
 where the F1-score is picked as the highest one via anomaly threshold selection. We also follow the same evaluation strategy by treating the whole window segment as correct if any observation in this segment is correctly detected as an anomaly \cite{Su2019}. Table \ref{tab:result} details the obtained performance results for our model as well as the three baselines. As one can see, our model keeps a great balance between precision and recall while the baselines tend to favor one of the metrics, and thus resulting to the best F1-score consistently over the baselines. On average, we achieve 17.8\% improvement over the best state-of-the-art performance.

 \subsection{Case Study}

Besides the performance gains FMUAD brings to the stage, another significant benefit of our model is its intuitiveness. In this section, we take a deep dive in two particular data samples FMUAD labelled as anomalous along with the model's output, to illustrate the detection mechanism.

Shown in Figure \ref{fig:case_study} are two cases: the left one is taken from MSL and the right one from SMD. In both cases, we sample a time period out of the time series containing anomalous labels, and the top chart (a1, a2) displays a hand-picked set of 6 metrics containing interesting and indicative patterns. Portions of the time series labelled as anomalies are highlighted in red, and the forecast error (anomaly score) output from the model is displayed in the lower chart (b1, b2).

In the first case study (a1, b1, MSL dataset), our model accurately made a capture on a temporal frequency pattern change. While the value range of the displayed metrics did not change in the anomaly event, metric 3 shows outputs a less-concentrated cluster of spikes, and metric 4 shows significantly more frequent ramps and fluctuations. While these are subtle patterns to human eyes, FMUAD correctly outputs a high anomaly score, indicating a high confidence in the anomaly judgement which matches the provided label.

In the second case study (a2, b2, SMD dataset), we demonstrate the model's capabilities of capturing value changes. The sample presented in (a2) contains two drastically different anomaly patterns: The anomaly on the left contains a short-lived, sparing spikes on metric channel 1. It is a prime example of a point anomaly. In contrast, the right anomaly has a longer duration, involves more metrics and contains a more subtle and sustained value increase. While it is easy to notice by humans, the statistical significance of the spatial pattern is only highly pronounced when aggregated over the entire period. Again, by design, FMUAD is able to detect both types of anomalies, evidently through a high forecasting error output.

Through the two case studies above, one can easily appreciate the intuitiveness of FMUAD. The model provides explainable and predictable outputs, opening up opportunities for further improvements and targeted fine-tuning based on the patterns of the target dataset. However, another major strength of the model remains to be addressed - its modularity. While we now have empirical evidence that FMUAD can pick up distinct patterns of anomaly, one naturally wonders if and how well the modularity provides a constructive impact, and whether each module indeed captures its target patterns.
 
  \begin{figure*}[!htbp]
    \centering
    \includegraphics[width=150mm]{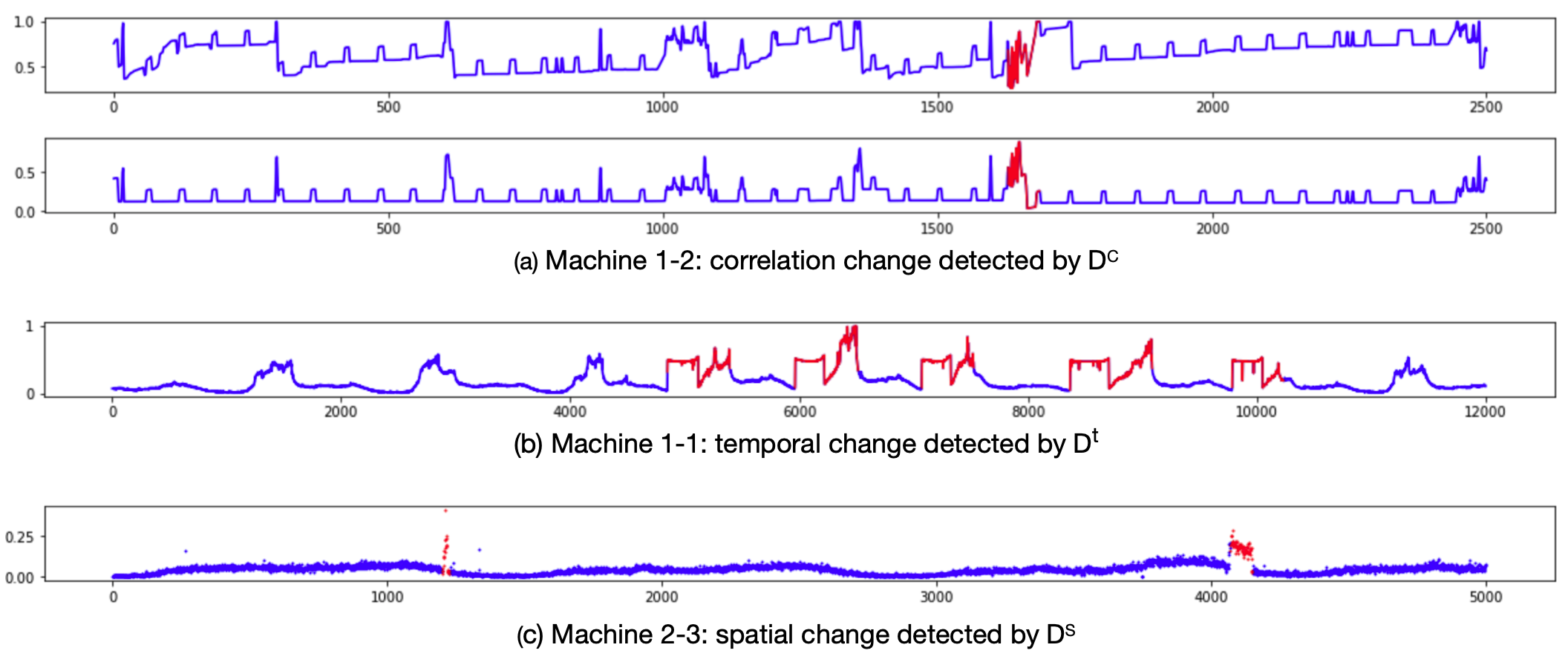}
    \caption{Anomaly patterns captured by different detectors.}
    \label{fig:ablation2}
\end{figure*}

 \subsection{Ablation Study}

To address the curiosity over the model's modularity, we perform an ablation study to analyze the effect of each detector module to the overall performance of the framework. To recap the model architecture, we use three separate and complementary forecast-based detector modules, each dedicated at one target pattern by design. Specifically, $D^c$ captures inter-series correlation, $D^t$ captures intra-series temporal patterns and $D^s$ captures multi-scale intra-series spatial patterns. We run the same benchmark experiments on all 28 machines of the SMD with only one of the three detectors enabled at a time, and compare the detection performance when that particular detector is present (and more importantly, when the other two are not present).

As with the case study, we sample a few time series each detector proves to perform well on, and present their plots in Figure \ref{fig:ablation2}. Table \ref{tab:detector_comparison} presents the F1-score for each of the detectors on each of the samples shown. The result again meets the expectation. In sample (a) where $D^c$ excels, we can see a clear change in the correlation: While in normal situations the two metrics are positively correlated, the labelled anomaly event depicts a negative correlation between the two time series. This is an instance where only $D^c$ is triggered, but not $D^s$ or $D^t$. Samples (b) and (c) general picture the same situation. In (b), a temporal change can be observed as the metric breaks out of its usual periodicity and manifests several sudden noisy, sudden peaks. These patterns are easily detected by $D^t$ but not the others. Finally, in (c), the metric takes a sharp point-anomaly on the left, and a more subtle but longer duration deviation on the right. Both deviations are registered in $D^s$, showing its strength in multi-scale spatial change detection.

\begin{table}[H]
\centering
\begin{tabular}{|c|c|c|c|}
\hline
Detector & Machine 1-1     & Machine 1-2     & Machine 2-3     \\ \hline
    $D^c$     & 0.6921          & \textbf{0.6976} & 0.0332          \\ \hline
    $D^t$     & \textbf{0.9984} & 0.5229          & 0.8968          \\ \hline
    $D^s$     & 0.9680          & 0.5884          & \textbf{0.9122} \\ \hline
\end{tabular}
\caption{Performance comparison (F1-score) among three detectors for representative machines from SMD.}
\label{tab:detector_comparison}
\end{table}
 
Of course, showing detection behavior on archetypal samples does not do justice to the modules' complementary effectiveness. As each detector works well for some cases, one may wonder whether the performance becomes diverse and unstable across datasets when enabling all three detectors. In the following Table \ref{tab:ablation2}, we present the model performance on each dataset when individual detectors are enabled alone, and when all three are in use. It is evident that combining the three detector modules provides a consistent best performance. More importantly, such combination makes our model performance so stable that only a 0.0001 variance in terms of F1-score remains cross the four set of datasets.
 
\begin{table}[H]
\begin{tabular}{|c|c|c|c|c|c|c|}
\hline
\multirow{2}{*}{} & \multicolumn{3}{c|}{SMD}    & \multirow{2}{*}{MSL} & \multirow{2}{*}{Mean} & \multirow{2}{*}{Variance} \\ \cline{2-4}
                  & group 1 & group 2 & group 3 &                      &                       &                           \\ \hline
             $D^c$     & 0.619   & 0.859   & 0.646   & 0.191                & 0.579                 & 0.0588                    \\ \hline
            $D^t$      & 0.586   & 0.866   & 0.731   & 0.851                & 0.759                 & 0.0127                    \\ \hline
             $D^s$     & 0.475   & 0.723   & 0.560   & 0.323                & 0.520                 & 0.0209                    \\ \hline
\textbf{FMUAD}             & 0.860   & 0.883   & 0.861   & 0.854                & \textbf{0.865}        & \textbf{0.0001}           \\ \hline
\end{tabular}
\caption{Module VS All - the overall framework outperforms each individual module consistently and exhibits excellent stability with only 0.0001 variance across different datasets.}
\label{tab:ablation2}
\end{table}

\begin{figure}[h]
    \centering
    \includegraphics[width=75mm]{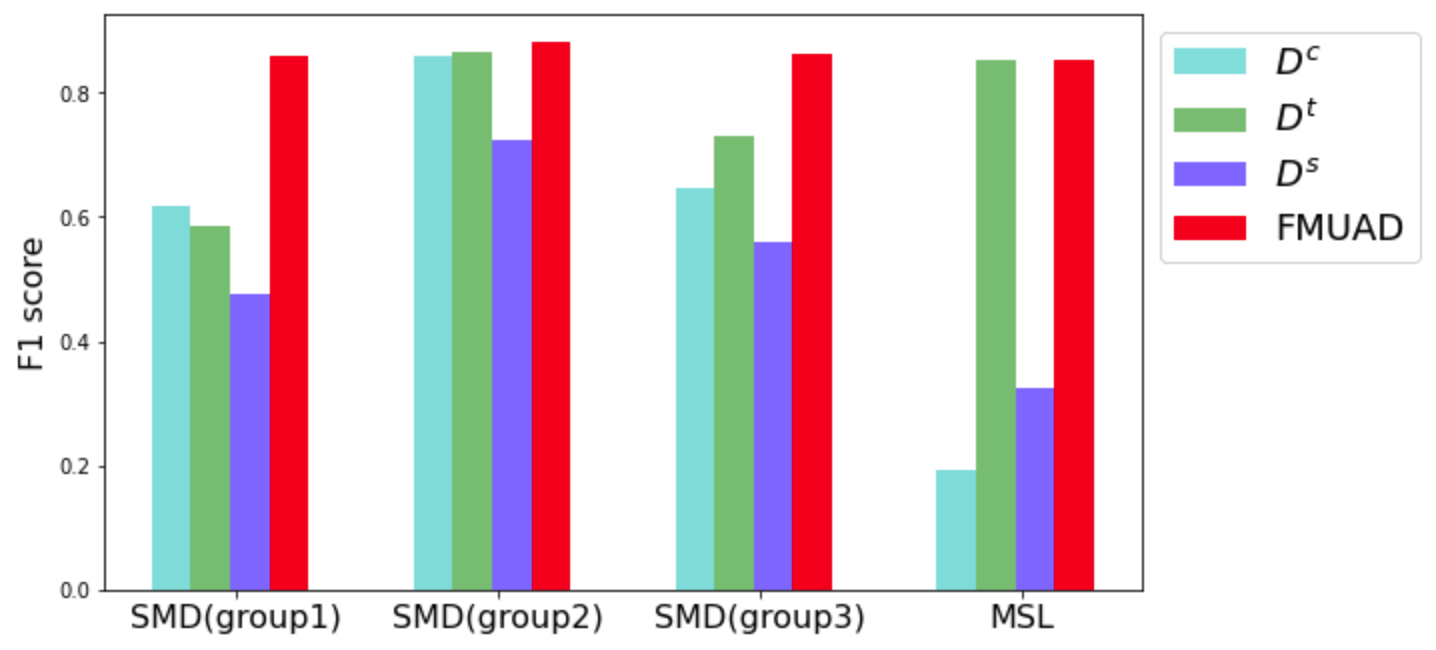}
    \caption{Model performance (F-1 score): Module VS All.}
    \label{fig:ablation}
\end{figure}

Besides the modular architecture, a novelty we would like to bring to attention is the batch variance term $\ell_2$ in the loss function. We incorporate this loss in the training time in hopes to learn a more stable and compact representation, therefore further improving the quality of anomaly detection. To test the extent of this improvement brought by $\ell_2$, we repeat the benchmark on each dataset but without the variance term in the loss function Eq. (\ref{loss_fun2}). The performance numbers are detailed in Table \ref{tab:ablation3}. Based on the numbers, it can be concluded that the variance control term has resulted in a consistent improvement in performance over when only $\ell_1$ is involved in training, albeit by a little. 

\begin{table}[h]
\centering
\begin{tabular}{|c|c|c|c|c|}
\hline
\multirow{2}{*}{}                                                          & \multicolumn{3}{c|}{SMD}    & \multirow{2}{*}{MSL} \\ \cline{2-4}
                                                                           & group 1 & group 2 & group 3 &                      \\ \hline
\begin{tabular}[c]{@{}c@{}}FMUAD w.o.\\ loss variance control \end{tabular} & 0.845   & 0.877   & 0.846   & 0.849                \\ \hline
\textbf{FMUAD}                                                             & \textbf{0.860}   & \textbf{0.883}   & \textbf{0.861}   & \textbf{0.854}                \\ \hline
\end{tabular}
\caption{Model performance (F1-score): Loss VS Loss without variance control.}
\label{tab:ablation3}
\end{table}

 \section{Conclusion}
 
In this paper, we discuss a Forecast-based Multi-aspect framework for Unsupervised multi-variate time series Anomaly Detection, or FMUAD in short. It is a bold attempt at this category of anomaly detection models by identifying the traits of anomaly patterns and categorizing them for a divide-and-conquer approach. We introduce a modular framework where the input time series is transformed into an intermediate representation better capturing each target pattern, and feeding it through each detector module that makes its own forecast on the said intermediate representation. We also experiment with a novel loss function that promotes compactness in the learned representation during training. To wrap them all up, we apply the loss function to compute the forecasting error, and thus produce an anomaly score.

We prove that the three detectors can indeed learn jointly for the optimal anomaly detection by contrasting its performance against state-of-the-art techniques on public reference datasets. The standard F-1 score from FMUAD is indeed higher than other models of its class. We also relate the detection mechanisms back to human intuition and find proof of each detector module performing to its advantage by agreeing to the anomaly patterns we assign them to detect. Furthermore, we investigate the role of modularity through an ablative study, and find that the combination of all three modules can achieve consistently higher and more stable performance than each one alone. We also prove that the compactness-focused $l_2$ loss term results in consistent improvement, albeit being minute compared to other improvements.

A recent trend we observe in the machine learning community is the tendency to move towards ``end-to-end" or ``one-size-fits-all" models, as a defiant departure from the old-school feature-engineering approaches. However, FMUAD may serve as a reminder that distinct patterns existing in the input data may still need a degree of tailored model design, all while being a self-adaptive model that remains end-to-end trainable.

This is clearly not the end of our journey, as the flexibility of FMUAD opens up a world of opportunities for further enhancements. For one, we can continue to discover new types of anomaly pattern and tailor a corresponding detector to its characteristics so those patterns are effectively captured; for two, there may exist better implementations for detectors $D^c$, $D^t$ and $D^s$ that can allow individual modules to perform even more optimally. Finally, the three detectors in FMUAD have operated in complete isolation of each other and the final aggregator. Enabling cross-interactions between these modules may also result in interesting observations that are worth investigating. All in all, we hope our model has availed additional potential for forecast-based anomaly detection models, and it can start a discussion to draw further work onto this general direction.

\bibliography{main} 
\bibliographystyle{ieeetr}

\end{document}